\documentclass[conference]{IEEEtran}

\ifdefined\pdfinfoomitdate
\fi
\ifdefined\pdfsuppressptexinfo
\fi

\usepackage{cite}

\usepackage[T1]{fontenc}
\usepackage[utf8]{inputenc}
\usepackage{graphicx}
\usepackage{booktabs}
\usepackage{array}
\usepackage{multirow}
\usepackage{tabularx}
\usepackage{amsmath}
\usepackage{amssymb}
\usepackage{xcolor}
\usepackage{url}
\usepackage{microtype}
\usepackage{float}

\graphicspath{{figures/}}
\title{Measuring and Improving Complex-Atomic Answer Consistency in Endoscopic VQA}

\author{\IEEEauthorblockN{Yuhao Liu, Cheng Zhao, Guanghui Yue}
\IEEEauthorblockA{School of Biomedical Engineering, Shenzhen University, Shenzhen, China}}

\begin{document}
\bstctlcite{IEEEexample:BSTcontrol}
\maketitle

\begin{abstract}
Endoscopic visual question answering (VQA) increasingly asks complex questions that combine several endoscopic answer components rather than isolated factual queries. Such complex answers may be scored as correct even when the same model fails on associated atomic questions. We introduce EndoCA, a paired complex-atomic answer consistency benchmark for evaluating whether complex answers remain consistent with same-image atomic answers. EndoCA contains two suites: EndoCA-Core evaluates compact question-complexity patterns commonly seen in practical endoscopic VQA, and EndoCA-Diagnostic supports controlled analysis across increasing question complexity. We evaluate 11 VLMs spanning open, medical, endoscopy-adapted, and closed-source models on EndoCA. Some VLMs achieve high complex-answer accuracy, yet their atomic-answer accuracy and complex-atomic answer consistency remain substantially lower. To reduce this complex-atomic inconsistency, we introduce Atomic-Support Reconciliation (ASR), a training-free mechanism that uses model-generated atomic answers as contextual premises for answer revision and consistency-guided selective answering. On four selected publicly available models, ASR-Revise improves paired complex-atomic correctness with modest changes in complex-answer accuracy, while ASR-Selective improves accuracy on answered cases by allowing the model to abstain from less reliable cases. Together, EndoCA and ASR provide a consistency-aware benchmark and a training-free mechanism for answer reconciliation and selective answering in endoscopic VQA.

\end{abstract}


\section{Introduction}
\label{sec:introduction}

Visual question answering (VQA) provides a task-directed way to evaluate visual understanding: a model must interpret an image under a natural-language query and return an answer grounded in the relevant visual evidence \cite{antol2015vqa}. Medical VQA extends this paradigm to clinically oriented image interpretation \cite{lin2023medicalvqa}. With stronger vision-language models (VLMs), endoscopic VQA is moving from isolated recognition prompts, such as whether a polyp is present, toward complex questions that combine multiple visual components. We use the term \emph{atomic question} to denote a same-image component check for one visual commitment, such as an anatomical site, a polyp count, or an instrument type. A complex question can combine several such checks into one natural answer.

\begin{figure}[!t]
    \centering
    \includegraphics[width=\linewidth,height=0.2\textheight]{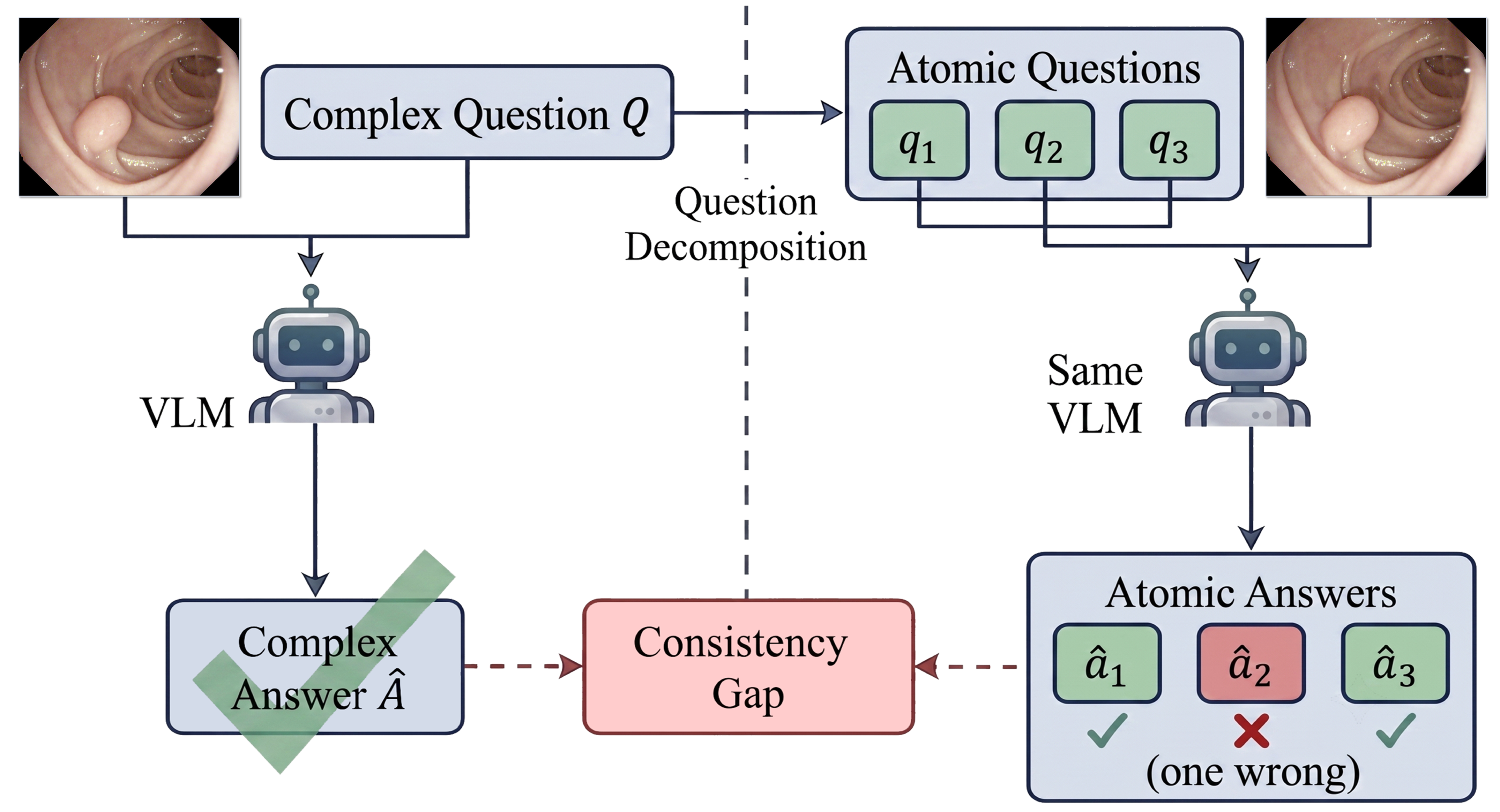}
    \caption{Paired complex-atomic evaluation. A correct complex answer can still disagree with associated atomic answers on the same image.}
    \label{fig:motivation}
\end{figure}

Recent endoscopic VLM benchmarks reflect this broader evaluation trend. For example, EndoBench evaluates models across multiple endoscopic scenarios, clinical tasks, and visual prompting granularities~\cite{liu2025endobench}. Such benchmarks provide important final-answer and task-performance signals. At the same time, when a complex question involves multiple component judgments, each component can be checked through a simpler atomic question. If the same model gives a correct complex answer but fails corresponding atomic questions under the same image and related question context, its image understanding and evidence use remain unstable in a way that complex-answer accuracy can obscure. The motivating example in Fig.~\ref{fig:motivation} illustrates this gap: the complex answer is accepted while one corresponding atomic answer is wrong. In medical settings, this hidden inconsistency matters because apparently correct answers may mask unreliable interpretation of the underlying endoscopic evidence.

We construct EndoCA from the two-level QA structure formed by Kvasir-VQA~\cite{gautam2024kvasirvqa} and Kvasir-VQA-x1~\cite{gautam2025kvasirvqa_x1}. Kvasir-VQA provides the component-level atomic QA pairs, while Kvasir-VQA-x1 combines corresponding same-image pairs into complex questions and naturalized complex answers. EndoCA keeps this link: each complex question is evaluated together with its associated atomic questions. EndoCA-Core covers common compact cases with one or two associated atomic questions and serves as the main comparison setting. EndoCA-Diagnostic follows the same protocol but adds a controlled three-atomic-question stratum to test how paired correctness changes as more endoscopic visual judgments are combined. Across both suites, EndoCA reports complex-answer accuracy, atomic-answer accuracy, joint accuracy, and complex-atomic inconsistency to separate final-answer correctness from same-image paired consistency.

Our study evaluates 11 VLMs from general open, medical open, endoscopy-adapted, and closed-source groups. Across both EndoCA suites, complex-answer rankings change once every associated atomic prediction must also be correct, and higher question complexity places stronger pressure on all-atomic accuracy. These results support reporting complex-answer accuracy together with atomic-answer accuracy, joint accuracy, and complex-atomic inconsistency whenever associated atomic questions are available. Building on this paired evaluation signal, we introduce Atomic-Support Reconciliation (ASR), a training-free mechanism that uses model-generated atomic answers as contextual premises for answer revision and consistency-guided selective answering.

This paper makes three contributions:
\begin{itemize}
    \item We introduce EndoCA, a paired complex-atomic answer consistency benchmark for endoscopic VQA, with EndoCA-Core for common compact question patterns and EndoCA-Diagnostic for controlled higher-complexity analysis.
    \item We evaluate 11 VLMs across general open, medical open, endoscopy-adapted, and closed-source groups, showing that complex-answer accuracy alone can hide substantial same-image complex-atomic answer inconsistency.
    \item We propose ASR, a training-free mechanism that uses model-generated atomic answers as contextual premises for answer revision and consistency-guided selective answering.
\end{itemize}

% Requires: graphicx, booktabs, array.
\begin{table*}[!t]
\centering
\scriptsize
\setlength{\tabcolsep}{2.05pt}
\setlength{\aboverulesep}{1pt}
\setlength{\belowrulesep}{1pt}
\renewcommand{\arraystretch}{1.00}
\caption{Representative EndoCA complex-atomic QA examples.}
\label{tab:endoca_examples}
\resizebox{0.955\textwidth}{!}{%
\begin{tabular}{>{\centering\arraybackslash}m{0.07\textwidth}>{\centering\arraybackslash}m{0.13\textwidth}>{\raggedright\arraybackslash}m{0.335\textwidth}>{\raggedright\arraybackslash}m{0.435\textwidth}}
\toprule
\textbf{\begin{tabular}[c]{@{}c@{}}Question-\\complexity\end{tabular}} & \textbf{Image} & \textbf{Complex QA} & \textbf{Atomic QA} \\
\midrule
\textbf{\large 1} & \includegraphics[width=0.78in,height=0.66in]{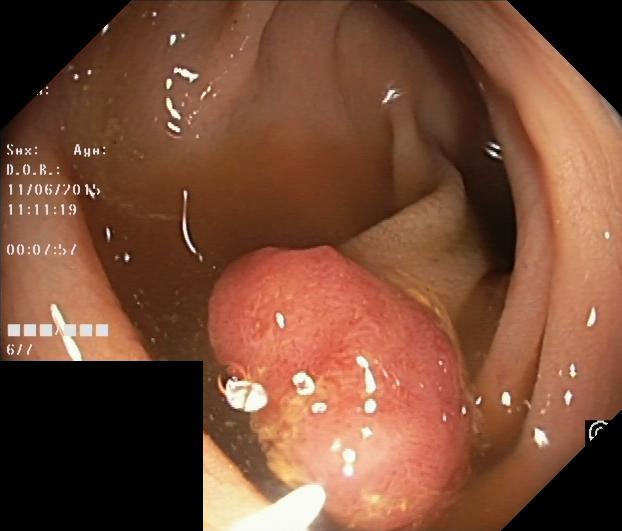} & \textbf{Q:} How many polyps are present in the gastrointestinal tract shown?\par\vspace{0.75mm}\textbf{A:} one polyp observed & \textbf{Polyp Count $\mathbf{q}_{\mathbf{1}}$:} How many polyps are in the image?\par\textbf{Atomic Answer $\mathbf{a}_{\mathbf{1}}$:} 1 \\
\midrule
\textbf{\large 2} & \includegraphics[width=0.78in,height=0.66in]{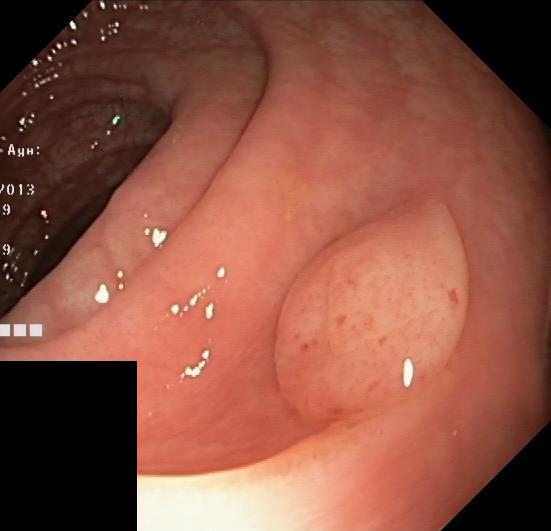} & \textbf{Q:} Are there any green or black box artifacts and is there visible text on the image?\par\vspace{0.75mm}\textbf{A:} Evidence of green and black box artifacts along with visible text is present. & \textbf{Box Artifact Presence $\mathbf{q}_{\mathbf{1}}$:} Is there a green/black box artifact?\par\textbf{Atomic Answer $\mathbf{a}_{\mathbf{1}}$:} yes\par\vspace{0.28mm}\textbf{Text Presence $\mathbf{q}_{\mathbf{2}}$:} Is there text?\par\textbf{Atomic Answer $\mathbf{a}_{\mathbf{2}}$:} yes \\
\midrule
\textbf{\large 3} & \includegraphics[width=0.78in,height=0.66in]{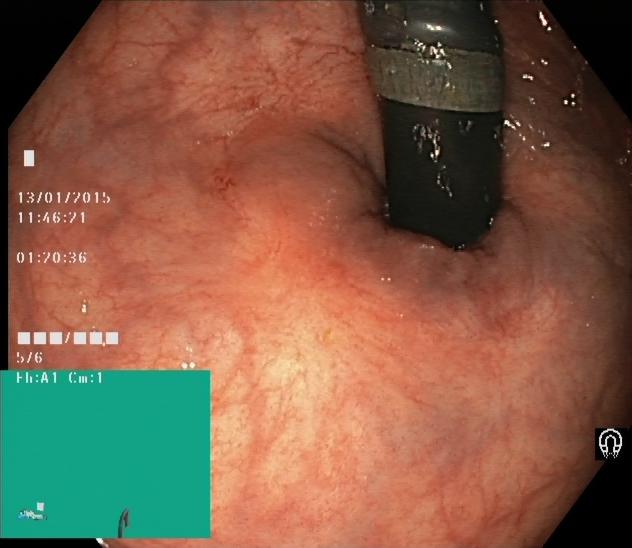} & \textbf{Q:} Where in the image are the instruments located, is there a green/black box artifact, and what instruments are visible?\par\vspace{0.75mm}\textbf{A:} A tube is visible in the center-right and upper-right regions, with a green/black box artifact present. & \textbf{Instrument Location $\mathbf{q}_{\mathbf{1}}$:} Where in the image is the instrument?\par\textbf{Atomic Answer $\mathbf{a}_{\mathbf{1}}$:} center-right; upper-right\par\vspace{0.28mm}\textbf{Box Artifact Presence $\mathbf{q}_{\mathbf{2}}$:} Is there a green/black box artifact?\par\textbf{Atomic Answer $\mathbf{a}_{\mathbf{2}}$:} yes\par\vspace{0.28mm}\textbf{Instrument Presence $\mathbf{q}_{\mathbf{3}}$:} Are there any instruments in the image? Check all that are present.\par\textbf{Atomic Answer $\mathbf{a}_{\mathbf{3}}$:} tube \\
\bottomrule
\end{tabular}%
}
\end{table*}

\section{Related Work}
\label{sec:related_work}

\subsection{Medical and Endoscopic VQA}
Medical VQA has progressed from modality-specific QA resources to broader clinical VLM benchmarks. VQA-RAD~\cite{lau2018vqarad}, SLAKE~\cite{liu2021slake}, and PathVQA~\cite{he2020pathvqa} provide foundations for radiology, semantically labeled medical QA, and pathology, respectively. In gastrointestinal endoscopy, Kvasir-VQA~\cite{gautam2024kvasirvqa} and Kvasir-VQA-x1~\cite{gautam2025kvasirvqa_x1} provide concise endoscopic QA and complex reasoning-oriented QA. Recent efforts further broaden the evaluation context: EndoBench evaluates VLMs across multiple endoscopic scenarios, clinical tasks, and visual prompting granularities~\cite{liu2025endobench}; Gastric-X studies multimodal gastric cancer analysis across VQA, report generation, retrieval, classification, and localization~\cite{lu2026gastricx}; and MEDVQA-GI explores challenge systems for gastrointestinal endoscopy VQA~\cite{gaihre2025medvqagi}. These studies expand the task scope and clinical context of medical and endoscopic VQA; their primary evaluation signals still center on final-answer accuracy and task performance, leaving model-internal consistency and stability less directly assessed.

\subsection{Consistency Evaluation and Answer Reconciliation}
Consistency-aware VQA and atomic evaluation study whether model answers remain reliable across related questions, structured reasoning steps, or smaller factual units. SQuINTing probes VQA models with lower-level perception sub-questions~\cite{selvaraju2020squinting}, and GQA provides structured real-world visual reasoning questions that support compositional evaluation~\cite{hudson2019gqa}. In generation evaluation, FActScore decomposes long-form text into atomic facts~\cite{min2023factscore}, while FaithScore and POPE assess VLM hallucination and image-grounded faithfulness at finer granularity~\cite{jing2024faithscore,li2023pope}. Reliability-oriented work has also appeared in surgical and endoscopic settings, such as SurgViVQA for temporally grounded surgical VideoQA robustness~\cite{drago2026surgvivqa}. EndoCA brings this consistency-aware atomic checking perspective to endoscopic VQA by focusing on whether accepted complex answers remain supported by their associated atomic answers.

Answer correction and self-checking methods use related predictions or multiple generations to improve reliability after an initial response. SOrT uses sub-question relevance to improve VQA consistency~\cite{dharur2021sorting}; self-consistency decoding samples multiple reasoning paths and selects an answer supported by agreement across paths~\cite{wang2023selfconsistency}; and SelfCheckGPT detects hallucination by comparing consistency across sampled passages~\cite{manakul2023selfcheckgpt}. In endoscopic diagnosis, EndoCogniAgent uses closed-loop self-consistency validation for agentic reasoning~\cite{tang2025endocogniagent}. ASR differs from prior self-checking methods by using model-generated atomic answers as explicit premises for complex-answer revision and as consistency evidence for selective answering.

\section{EndoCA Benchmark}
\label{sec:endoca}

\subsection{Source Dataset}
EndoCA uses the paired QA structure of Kvasir-VQA and Kvasir-VQA-x1. Kvasir-VQA contains concise QA pairs over gastrointestinal endoscopic images, covering common endoscopic findings, landmarks, instruments, and artifacts. Kvasir-VQA-x1 records complex questions and naturalized complex answers formed from corresponding same-image QA pairs. In EndoCA, the associated atomic questions are these same-image component questions linked to a complex question. This recorded association links each complex question to its atomic answer annotations, providing the paired units used for complex-atomic evaluation.

\subsection{Benchmark Construction}
The construction process starts from samples that contain an endoscopic image, a non-empty complex question, a complex answer annotation, and at least one associated atomic question-answer pair. Fixed parsing, answer normalization, consistency checks, and validity filtering standardize QA fields, canonicalize answer forms, require complete image/question/answer fields, well-formed atomic-question sets, and supported question types, and verify agreement between the complex annotation and its associated atomic annotations under the same scoring rules. The resulting paired units are illustrated in Table~\ref{tab:endoca_examples}.

EndoCA is organized into two complementary suites. Question complexity is the number of associated atomic questions in a complex answer. EndoCA-Core is the main evaluation suite, with 12,000 complex samples, 15,736 atomic QA items, and 27,736 total QA items per model; it focuses on complexity-1 and complexity-2 cases for stable cross-model comparison. EndoCA-Diagnostic contains 6,000 complex samples, 9,300 atomic QA items, and 15,300 total QA items per model; it keeps compact questions as the majority and adds 900 complexity-3 cases as a controlled higher-complexity stratum. The benchmark atomic questions cover high-confidence endoscopic question types over common findings, landmarks, instruments, and artifacts. The suites use distinct sample identities: Core provides the main benchmark distribution, and Diagnostic provides a separate setting for question-complexity analysis.

\subsection{Answer Scoring and Metrics}
For each sample \(i\), EndoCA stores an endoscopic image \(V_i\), a complex question \(Q_i\), a complex answer annotation \(A_i\), and an associated atomic-question list \(U_i\):
\begin{equation}
\begin{aligned}
x_i&=(V_i,Q_i,A_i,U_i),\\
U_i&=((q_{i,1},a_{i,1}),\ldots,(q_{i,m_i},a_{i,m_i})).
\end{aligned}
\end{equation}
where \(q_{ij}\) and \(a_{ij}\) are the \(j\)-th associated atomic question and its answer annotation. The quantity \(m_i=|U_i|\) is the question complexity, defined as the number of atomic questions associated with \(Q_i\).

EndoCA evaluates answers with a fixed question-type-aware scorer. The same concise-answer prompt template is used for complex and atomic probes, with the question field replaced by \(Q_i\) or \(q_{ij}\). Model outputs may be short phrases or complete sentences, so the scorer normalizes common surface variation, including case, spacing, punctuation, yes/no forms, count words, and canonical labels. For separate atomic predictions, the expected question type determines the matching rule: binary questions are mapped to yes/no values, count questions to integers, categorical questions to canonical labels, and multi-label questions to normalized label sets. For complex-answer scoring, EndoCA uses the verified associated atomic annotations as component-level ground truth for the original complex question. A complex prediction is counted as correct only when the normalized answer matches every associated atomic answer annotation under the corresponding question-type rule. Empty outputs, execution errors, ambiguous answers, and non-committal responses are counted as incorrect. This fixed protocol enables reproducible comparison across open-source and closed-source VLMs and keeps the evaluation independent of an additional LLM judge.

\begin{table}[!t]
\centering
\caption{Example scoring of an EndoCA paired unit.}
\label{tab:scoring_example}
\scriptsize
\setlength{\tabcolsep}{2.4pt}
\renewcommand{\arraystretch}{1.06}
\begin{tabularx}{\columnwidth}{@{}>{\raggedright\arraybackslash}p{0.19\columnwidth}>{\raggedright\arraybackslash}X@{}}
\toprule
\textbf{Step} & \textbf{Scoring process and result} \\
\midrule
Paired unit &
\(x_i=(V_i,Q_i,A_i,U_i)\), where \(U_i\) contains two associated atomic QA annotations with \(a_{i,1}=\) yes and \(a_{i,2}=\) yes. \\
\addlinespace[0.5mm]
Prompt template &
Complex and atomic probes use the same template, replacing only \(\langle Q\rangle\):
\emph{Answer the question based on the endoscopic image. Respond with a concise short answer only. Do not explain. Question: \(\langle Q\rangle\)} \\
\addlinespace[0.5mm]
Complex QA &
\(\langle Q\rangle=Q_i\): \emph{Are there any green/black box artifacts and is there visible text in the image?}
Raw output \(\hat{A}_i\): \emph{A green/black box artifact is present, and visible text is present.} \\
\addlinespace[0.5mm]
Complex score &
The scorer extracts component labels: box artifact = yes and visible text = yes. Both match \((a_{i,1},a_{i,2})\), so \(\mathcal{C}_i=1\). \\
\addlinespace[0.5mm]
Atomic QA &
\(\langle Q\rangle=q_{i,1}\): \emph{Is there a green/black box artifact?}
Raw output \(\hat{a}_{i,1}=\) \emph{yes}.
\(\langle Q\rangle=q_{i,2}\): \emph{Is there visible text?}
\(\hat{a}_{i,2}=\) \emph{no}. \\
\addlinespace[0.5mm]
Atomic score &
\(\hat{a}_{i,1}\) matches \(a_{i,1}\), so \(\mathcal{A}_{i,1}=1\).
\(\hat{a}_{i,2}\) does not match \(a_{i,2}\), so \(\mathcal{A}_{i,2}=0\). \\
\addlinespace[0.5mm]
Indicators &
After normalization, \(\mathcal{C}_i=1\), \(\mathcal{A}_{i,1}=1\), \(\mathcal{A}_{i,2}=0\), and \(\mathcal{A}^{\star}_i=0\). \\
\bottomrule
\end{tabularx}
\end{table}

For a model output, let \(\hat{A}_i\) be the complex answer and \(\hat{a}_{ij}\) be the answer to atomic question \(q_{ij}\). Let \(S_C(\hat{A}_i,x_i)\in\{0,1\}\) denote complex-answer correctness, and \(S_A(\hat{a}_{ij},q_{ij},a_{ij})\in\{0,1\}\) denote atomic-answer correctness. We define
\begin{equation}
\label{eq:correctness_indicators}
\begin{aligned}
\mathcal{C}_i &= S_C(\hat{A}_i,x_i), &
\mathcal{A}_{ij} &= S_A(\hat{a}_{ij},q_{ij},a_{ij}),\\
\mathcal{A}^{\star}_i &= \prod_{j=1}^{m_i} \mathcal{A}_{ij}.
\end{aligned}
\end{equation}
Here, \(\mathcal{C}_i=1\) when the complex prediction matches all associated atomic answer annotations for sample \(i\), \(\mathcal{A}_{ij}=1\) when the model's answer to \(q_{ij}\) matches its associated annotation, and \(\mathcal{A}^{\star}_i=1\) only when all atomic questions for the sample are answered correctly. In question-complexity analysis, all-atomic accuracy averages \(\mathcal{A}^{\star}_i\) within each complexity group.

EndoCA reports four metrics. Complex-answer accuracy measures the accuracy of model answers to complex questions. Atomic-answer accuracy averages correctness across the model's separate answers to the associated atomic questions. Joint accuracy requires both the complex prediction and every separately generated atomic prediction to match their associated answer annotations for the same sample. Complex-atomic inconsistency is computed among correct complex-answer cases and measures the share with at least one wrong atomic prediction:
\begin{align}
\text{Complex-answer Accuracy} &=
\frac{1}{N}\sum_{i=1}^{N} \mathcal{C}_i,\\
\text{Atomic-answer Accuracy} &=
\frac{\sum_{i=1}^{N}\sum_{j=1}^{m_i} \mathcal{A}_{ij}}
{\sum_{i=1}^{N} m_i},\\
\text{Joint Accuracy} &=
\frac{1}{N}\sum_{i=1}^{N}
\mathcal{C}_i\mathcal{A}^{\star}_i,\\
\text{Complex-Atomic Inconsistency} &=
\frac{\sum_{i=1}^{N}\mathcal{C}_i(1-\mathcal{A}^{\star}_i)}
{\sum_{i=1}^{N}\mathcal{C}_i}.
\end{align}
If a model has no correct complex-answer cases, this conditional metric is undefined and should be reported as N/A; in our evaluated rows, every model has at least one correct complex answer, so the denominator is nonzero. Result tables abbreviate the three accuracy metrics as Complex Acc., Atomic Acc., and Joint Acc.; the inconsistency metric is written out.

\section{Method}
\label{sec:method}

\subsection{Motivation}
EndoCA exposes complex-atomic inconsistency in endoscopic VQA: a model may give an acceptable complex answer while failing one supporting same-image atomic question. This observation suggests a direct inference-time mitigation strategy. ASR uses only the image, the complex question, the direct model answer, and model-generated atomic answers; the associated atomic annotations are used only for EndoCA scoring. Atomic-Support Reconciliation (ASR) treats the model-generated atomic answers as explicit component-level evidence for answer revision and selective answering.

\begin{figure}[H]
\centering
\fbox{\begin{minipage}{0.94\linewidth}
\footnotesize
\textbf{Workflow 1: Atomic-Support Reconciliation}\\
\textbf{Input:} image \(V_i\), complex question \(Q_i\), associated atomic questions \(\{q_{ij}\}\), VLM \(M\).\\
\textbf{1. Direct answer:} \(\hat{A}_i \leftarrow M(V_i,Q_i)\).\\
\textbf{2. Atomic answers:} \textbf{for} \(j=1,\ldots,m_i\) \textbf{do}\\
\hspace*{3mm}\(\hat{a}_{ij}\leftarrow M(V_i,q_{ij})\).\\
\textbf{end for}\\
\textbf{3. Atomic premises:} \(P_i\leftarrow\{(q_{ij},\hat{a}_{ij})\}_{j=1}^{m_i}\).\\
\textbf{4. Revise:} \(\tilde{A}_i\leftarrow M(V_i,Q_i,\hat{A}_i,P_i)\).\\
\textbf{5. Select:} \(d_i\leftarrow M(V_i,Q_i,\tilde{A}_i,P_i)\),\\
\hspace*{3mm}checking consistency between \(\tilde{A}_i\) and \(P_i\),\\
\hspace*{3mm}where \(d_i\in\{\text{answer},\text{abstain}\}\).\\
\textbf{Return:}\\[-0.5mm]
\hspace*{3mm}ASR-Revise: \(\tilde{A}_i\).\\[-0.2mm]
\hspace*{3mm}ASR-Selective: \(\tilde{A}_i\) if \(d_i=\text{answer}\), otherwise abstain.
\end{minipage}}
\caption{ASR workflow for revision and selective answering.}
\label{fig:asr_workflow}
\end{figure}

\subsection{ASR-Revise}
ASR-Revise revises complex answers with atomic answers through a training-free answer reconciliation procedure.
For each sample, the model first produces a direct complex answer \(\hat{A}_i\) to the complex question \(Q_i\). The same model then answers every associated atomic question \(q_{ij}\), yielding \(\hat{a}_{ij}\). ASR organizes the atomic question-answer pairs into an atomic premise set \(P_i\), which records the model's own same-image component-level evidence:
\begin{equation}
P_i=\{(q_{ij},\hat{a}_{ij})\}_{j=1}^{m_i}.
\end{equation}

The revise step asks the model to produce a reconciled complex answer \(\tilde{A}_i\) using the image, the complex question, the model's direct complex answer, and the model-generated atomic answers as contextual premises. This preserves the original complex-answer task while exposing the model to a structured summary of its atomic responses. All ASR calls use a fixed prompt template per variant: ASR-Revise requests a concise revised answer, and ASR-Selective requests an explicit answer/abstain decision together with the revised answer and a coarse consistency label. Selective-answering coverage counts only samples with an explicit answer decision; revised text is scored by the same EndoCA protocol as direct complex answers. By conditioning on the model-generated atomic answers and applying a consistency-based answer decision, ASR can improve joint accuracy and reduce complex-atomic inconsistency even when full-coverage complex-answer accuracy changes only modestly.

\subsection{ASR-Selective}
ASR-Selective performs selective answering under complex-atomic consistency. After ASR-Revise produces \(\tilde{A}_i\), the model is asked to judge whether the revised complex answer is consistent with the atomic premise set \(P_i\). The output is a compact decision object containing an ``answer''/``abstain'' decision, the revised answer, and a coarse consistency label. If the decision is ``answer'', ASR-Selective returns \(\tilde{A}_i\); otherwise, it abstains for that sample.

This selective-answering step provides a controlled way to withhold low-consistency responses. We evaluate it with coverage, the percentage of samples for which an answer is returned, and selective accuracy, the complex-answer accuracy on the answered subset. In endoscopic VQA, this reliability--coverage view is useful because the model returns more answers when its complex answer and model-generated atomic answers agree, and fewer answers when the paired evidence is unstable.

\section{Experiments}
\label{sec:experiments}
\begin{table*}[!t]
\centering
\begin{minipage}[t]{0.485\textwidth}
\centering
\scriptsize
\setlength{\tabcolsep}{1.35pt}
\renewcommand{\arraystretch}{1.03}
\caption{EndoCA-Core paired evaluation.}
\label{tab:core_results}
\resizebox{\linewidth}{!}{%
\begin{tabular}{@{}llcrrrrr@{}}
\toprule
Group & Model & Params & \shortstack{QA\\Count} & \shortstack{Complex\\Acc. (\%)} & \shortstack{Atomic\\Acc. (\%)} & \shortstack{Joint\\Acc. (\%)} & \shortstack{Complex-Atomic\\Inconsistency (\%)} \\
\midrule
\multirow{3}{*}{\shortstack[l]{General\\open}} & InternVL2.5-8B & 8B & 27,736 & 51.4 & 44.0 & 31.5 & 38.7 \\
& Qwen2.5-VL-7B & 7B & 27,736 & 54.1 & 55.5 & 38.0 & 29.8 \\
& Qwen3-VL-8B & 8B & 27,736 & 55.1 & 67.3 & 47.6 & 13.7 \\
\midrule
\multirow{3}{*}{\shortstack[l]{Medical\\open}} & MedGemma-4B & 4B & 27,736 & 35.4 & 29.1 & 17.1 & 51.8 \\
& Lingshu-7B & 7B & 27,736 & 51.6 & 53.7 & 36.9 & 28.6 \\
& LLaVA-Med & 7B & 27,736 & 28.7 & 17.0 & 4.9 & 83.0 \\
\midrule
\multirow{2}{*}{\shortstack[l]{Endoscopy\\-adapted}} & MedGemma-FT & 4B & 27,736 & \textbf{74.5} & 73.0 & 55.8 & 25.1 \\
& Qwen2.5-VL-FT & 7B & 27,736 & 72.1 & 59.3 & 41.6 & 42.4 \\
\midrule
\multirow{3}{*}{\shortstack[l]{Closed\\source}} & GPT-5.5 & -- & 27,736 & 60.8 & 63.0 & 47.4 & 22.1 \\
& Grok-4.20 & -- & 27,736 & 34.8 & 36.9 & 19.7 & 43.3 \\
& Claude Opus 4.7 & -- & 27,736 & 63.9 & \textbf{76.1} & \textbf{56.0} & \textbf{12.5} \\
\bottomrule
\end{tabular}%
}
\end{minipage}
\hfill
\begin{minipage}[t]{0.485\textwidth}
\centering
\scriptsize
\setlength{\tabcolsep}{1.6pt}
\renewcommand{\arraystretch}{1.03}
\caption{EndoCA-Diagnostic paired evaluation.}
\label{tab:diagnostic_results}
\resizebox{\linewidth}{!}{%
\begin{tabular}{@{}llcrrrrr@{}}
\toprule
Group & Model & Params & \shortstack{QA\\Count} & \shortstack{Complex\\Acc. (\%)} & \shortstack{Atomic\\Acc. (\%)} & \shortstack{Joint\\Acc. (\%)} & \shortstack{Complex-Atomic\\Inconsistency (\%)} \\
\midrule
\multirow{3}{*}{\shortstack[l]{General\\open}} & InternVL2.5-8B & 8B & 15,300 & 46.1 & 43.6 & 26.7 & 42.0 \\
& Qwen2.5-VL-7B & 7B & 15,300 & 48.1 & 55.0 & 33.0 & 31.4 \\
& Qwen3-VL-8B & 8B & 15,300 & 50.3 & 66.0 & 42.1 & \textbf{16.3} \\
\midrule
\multirow{3}{*}{\shortstack[l]{Medical\\open}} & MedGemma-4B & 4B & 15,300 & 32.1 & 28.4 & 14.8 & 54.0 \\
& Lingshu-7B & 7B & 15,300 & 46.3 & 54.1 & 32.9 & 28.9 \\
& LLaVA-Med & 7B & 15,300 & 22.6 & 15.3 & 3.8 & 83.1 \\
\midrule
\multirow{2}{*}{\shortstack[l]{Endoscopy\\-adapted}} & MedGemma-FT & 4B & 15,300 & 70.4 & \textbf{71.2} & \textbf{50.7} & 28.0 \\
& Qwen2.5-VL-FT & 7B & 15,300 & \textbf{70.5} & 58.0 & 36.8 & 47.8 \\
\midrule
\multirow{3}{*}{\shortstack[l]{Closed\\source}} & GPT-5.5 & -- & 15,300 & 55.2 & 61.0 & 41.7 & 24.4 \\
& Grok-4.20 & -- & 15,300 & 30.1 & 35.8 & 16.7 & 44.6 \\
& Claude Opus 4.7 & -- & 15,300 & 54.2 & 56.6 & 36.0 & 33.5 \\
\bottomrule
\end{tabular}%
}
\end{minipage}
\end{table*}

\subsection{Experimental Setup}

EndoCA evaluates whether a model's complex prediction matches the associated atomic answer annotations and remains consistent with the model's separate answers to the associated atomic questions on the same image. We evaluate 11 VLMs from four model groups. The General open group includes InternVL2.5-8B~\cite{chen2024internvl25}, Qwen2.5-VL-7B~\cite{bai2025qwen25vl}, and Qwen3-VL-8B~\cite{bai2025qwen3vl}. The Medical open group includes MedGemma-4B~\cite{google2025medgemma}, Lingshu-7B~\cite{lingshu2025modelcard}, and LLaVA-Med~\cite{li2023llavamed}. The endoscopy-adapted group includes MedGemma-FT and Qwen2.5-VL-FT~\cite{gautam2025kvasirvqa_x1}, which are LoRA-adapted on the endoscopic VQA domain~\cite{hu2022lora}. We use them as in-domain adaptation references: they are expected to be strong on complex-answer accuracy, and the analysis asks whether that strength also carries over to paired atomic correctness. The Closed-source group includes GPT-5.5~\cite{openai2026models}, Grok-4.20~\cite{xai2026grok}, and Claude Opus 4.7~\cite{anthropic2026claude}. Each model answers every complex question and every associated atomic question on the same image, and all outputs are scored with the fixed question-type-aware scorer defined in Section~\ref{sec:endoca}.

All local open-source models are evaluated with PyTorch on a server with 8 NVIDIA RTX 4090 GPUs. All prompts are zero-shot and contain one endoscopic image, one user question, and the fixed concise-answer instruction described in Section~\ref{sec:endoca}, with no system prompt or in-context examples. Open-source models use deterministic decoding with \texttt{do\_sample=False} and a maximum generation length of 80 tokens. Closed-source models are evaluated with \texttt{temperature=0} and \texttt{max\_tokens=80}; their reasoning settings are set to \texttt{xhigh} for GPT-5.5 and Grok-4.20, and \texttt{max} for Claude Opus 4.7. Images are loaded in RGB format and processed by each model's default visual processor, without manual cropping or global resizing; closed-source API calls use base64-encoded original images. All reported runs cover the expected QA Count, and empty or invalid outputs are scored as incorrect under the fixed EndoCA scoring protocol.

\subsection{EndoCA Benchmark Results}
On EndoCA-Core, complex-answer accuracy and paired atomic correctness diverge substantially (Table~\ref{tab:core_results}). The in-domain adapted references achieve the highest complex-answer accuracy, with MedGemma-FT and Qwen2.5-VL-FT reaching 74.5\% and 72.1\%, respectively. Their joint accuracy differs by 14.2 percentage points, indicating that similar complex-answer performance can correspond to different levels of atomic-answer correctness. Across open and closed-source models, rankings based on complex-answer accuracy change when every separately generated atomic prediction must also be correct.
This supports the motivating concern that an accepted endoscopic complex answer may still hide unsupported component judgments.

\begin{figure}[!t]
    \centering
    \includegraphics[width=\linewidth]{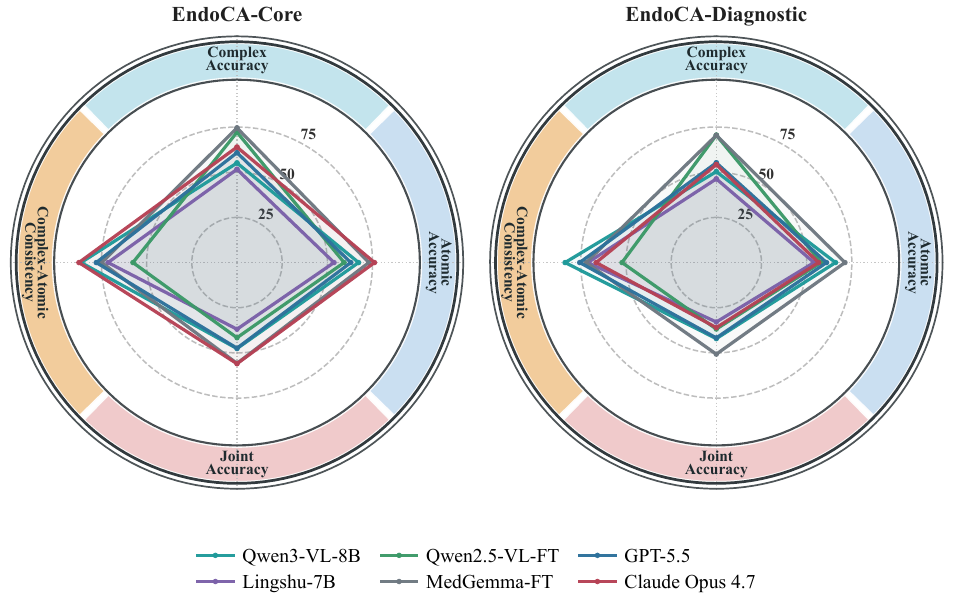}
    \caption{Accuracy--consistency radar chart for six representative VLMs on EndoCA-Core and EndoCA-Diagnostic. Complex-atomic consistency denotes the complement of inconsistency.}
    \label{fig:radar_chart}
\end{figure}

The diagnostic suite confirms the same reliability gap under a controlled complexity axis. MedGemma-FT and Qwen2.5-VL-FT obtain nearly identical complex-answer accuracy, 70.4\% and 70.5\%, yet their joint accuracy differs substantially, 50.7\% versus 36.8\% (Table~\ref{tab:diagnostic_results}). Their complex-atomic inconsistency also differs by 19.8 percentage points. This contrast is central to EndoCA: paired metrics expose whether a correct complex response remains aligned with the model's same-image atomic predictions.

The six-model radar summary reinforces this separation between accepted complex answers and paired atomic correctness. Several models remain competitive on complex-answer accuracy but drop on joint accuracy or complex-atomic consistency. For visualization, Fig.~\ref{fig:radar_chart} plots the complement of complex-atomic inconsistency so that all axes follow a higher-is-better direction.

\begin{figure}[!t]
    \centering
    \includegraphics[width=\linewidth]{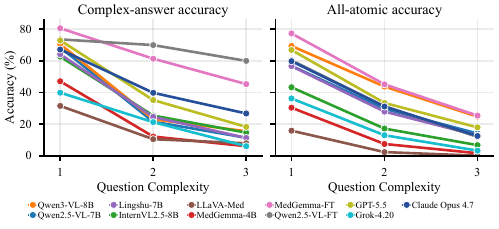}
    \caption{Question-complexity analysis on EndoCA-Diagnostic. All-atomic accuracy requires every separate atomic prediction to match its associated answer annotation.}
    \label{fig:question_complexity}
\end{figure}

Question complexity makes the same effect visible at the sample level. All-atomic accuracy uses \(\mathcal{A}^{\star}_i\) from Eq.~\eqref{eq:correctness_indicators}; a single wrong atomic prediction sets the sample-level indicator to zero. At complexity 3, several models retain non-trivial complex-answer accuracy while all-atomic accuracy remains much lower (Fig.~\ref{fig:question_complexity}). As a complex question combines more atomic judgments, all-atomic accuracy becomes a paired correctness stress test.

\subsection{Evaluation of ASR}
We next evaluate whether ASR can reduce complex-atomic inconsistency by using model-generated atomic answers as contextual premises. We apply ASR to four stronger publicly available models selected from the general open, medical open, and endoscopy-adapted groups: Qwen3-VL-8B, Lingshu-7B, MedGemma-FT, and Qwen2.5-VL-FT. Relative to the direct EndoCA-Core results, ASR-Revise improves joint accuracy for all four models, with gains from +3.1 to +8.1 percentage points, and reduces complex-atomic inconsistency by 10.4 points on average (Table~\ref{tab:asr_results}). Its effect on full-coverage complex-answer accuracy is mixed, with the clearest benefit appearing in paired reconciliation.

\begin{table}[H]
\centering
\scriptsize
\setlength{\tabcolsep}{1.8pt}
\renewcommand{\arraystretch}{1.04}
\caption{ASR-Revise and ASR-Selective results on EndoCA-Core.}
\label{tab:asr_results}
\begin{tabular}{@{}lrr@{\hspace{3pt}}c@{\hspace{4pt}}c@{\hspace{3pt}}rr@{}}
\toprule
Model & {\scriptsize\shortstack{Direct\\Complex\\Acc. (\%)}} & {\scriptsize\shortstack{Revise\\Complex\\Acc. (\%)}} & {\scriptsize\shortstack{Joint\\Acc.\\\(\Delta\)}} & {\scriptsize\shortstack{Complex-Atomic\\Inconsistency\\\(\Delta\)}} & {\scriptsize\shortstack{Coverage\\(\%)}} & {\scriptsize\shortstack{Selective\\Complex\\Acc. (\%)}} \\
\midrule
Qwen3-VL-8B & 55.1 & 54.7 & +5.3 & -10.3 & 73.0 & 68.3 \\
MedGemma-FT & 74.5 & 76.0 & +8.1 & -9.2 & 95.7 & 76.8 \\
Qwen2.5-VL-FT & 72.1 & 69.6 & +3.1 & -6.6 & 76.1 & 74.9 \\
Lingshu-7B & 51.6 & 48.4 & +5.2 & -15.5 & 58.9 & 63.9 \\
\midrule
Average & 63.3 & 62.2 & +5.4 & -10.4 & 75.9 & 71.0 \\
\bottomrule
\end{tabular}
\end{table}

\begin{figure}[!b]
    \centering
    \includegraphics[width=\linewidth]{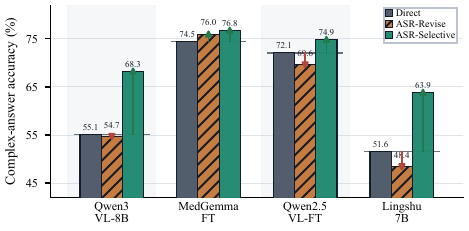}
    \caption{Direct, ASR-Revise, and ASR-Selective complex-answer accuracy on EndoCA-Core. Direct and ASR-Revise accuracies use all samples; ASR-Selective accuracy is computed over answered samples only.}
    \label{fig:asr_core}
\end{figure}

ASR-Selective characterizes a reliability--coverage trade-off for the answered subset. On average, it returns answers for 75.9\% of samples and reaches 71.0\% selective complex-answer accuracy. Models with less stable paired behavior improve answered-subset accuracy by abstaining more often. Fig.~\ref{fig:asr_core} compares Direct, ASR-Revise, and ASR-Selective complex-answer accuracy together with the coverage reported in Table~\ref{tab:asr_results}.

\subsection{Ablation Study}
\label{sec:ablation_atomic_premises}

We isolate the role of atomic premises using no-atomic self-check controls on the same four models and EndoCA-Core samples. Direct is the original complex answer. For Complex Acc., Joint Acc., and Complex-Atomic Inconsistency, w/o atomic is Self-Revise, which revises using the image, complex question, and direct answer without model-generated atomic premises. For Coverage and Selective Complex Acc., w/o atomic is Self-Selective, which makes the answer/abstain decision without atomic consistency evidence. ASR adds atomic premises to the corresponding revise and selective variants.

\begin{table}[H]
\centering
\scriptsize
\setlength{\tabcolsep}{2.6pt}
\renewcommand{\arraystretch}{1.05}
\caption{Atomic-premise ablation on EndoCA-Core.}
\label{tab:no_atomic_ablation_compact}
\resizebox{\columnwidth}{!}{%
\begin{tabular}{@{}lrrrrr@{}}
\toprule
Method & \shortstack{Complex\\Acc. (\%)} & \shortstack{Joint\\Acc. (\%)} & \shortstack{Complex-Atomic\\Inconsistency (\%)} & \shortstack{Coverage\\(\%)} & \shortstack{Selective Complex\\Acc. (\%)} \\
\midrule
Direct & \textbf{63.3} & 45.5 & 27.4 & 100.0 & 63.3 \\
w/o atomic & 60.6 & 43.0 & 28.2 & 84.5 & 66.8 \\
ASR & 62.2 & \textbf{50.9} & \textbf{17.1} & 75.9 & \textbf{71.0} \\
\bottomrule
\end{tabular}%
}
\end{table}

The ablation shows that generic self-checking alone does not recover the paired-correctness gains of ASR. Self-Revise without atomic premises reduces full-coverage complex-answer accuracy from 63.3\% to 60.6\% and does not improve joint accuracy or inconsistency, whereas ASR keeps complex-answer accuracy close to Direct at 62.2\%, improves joint accuracy to 50.9\%, and reduces complex-atomic inconsistency to 17.1\% (Table~\ref{tab:no_atomic_ablation_compact}). The selective columns show a related pattern: Self-Selective filters some unreliable answers, reaching 66.8\% answered-subset accuracy at 84.5\% coverage, while ASR-Selective achieves 71.0\% at 75.9\% coverage by using atomic premises as a stronger consistency signal. These results indicate that conditioning revision and selection on model-generated atomic answers is the key source of the paired-correctness gains.

\section{Conclusion}
\label{sec:conclusion}

This paper presented EndoCA, a paired complex-atomic evaluation protocol for endoscopic VQA that scores complex predictions together with same-image atomic predictions. Across 11 VLMs, complex-answer accuracy gives an incomplete view of paired correctness: models with strong complex-answer performance can still show lower atomic-answer accuracy, lower joint accuracy, and non-negligible complex-atomic inconsistency. ASR further shows that model-generated atomic answers can serve as contextual premises for answer revision and consistency-guided selective answering, exposing a reliability--coverage trade-off. The fixed EndoCA scorer enables reproducible comparison across model groups, while future work should extend paired evaluation to free-form clinical questions through reliable automatic decomposition and careful comparison of decomposition strategies.

\bibliographystyle{IEEEtran}
\bibliography{references}

\end{document}